%% file: main.tex
\documentclass[10pt]{article} 
\usepackage[preprint]{tmlr}

\input{math_commands.tex}

\usepackage{hyperref}
\usepackage{url}
\usepackage{graphicx}
\usepackage{booktabs}
\usepackage[table]{xcolor}
\usepackage[dvipsnames]{xcolor}
\usepackage{lmodern}

\newcommand{\leopard}{$\overline{\text{L}}$EOPARD }

\newcommand{\redhl}[1]{\colorbox{RedOrange}{\textit{#1}}}
\newcommand{\greenhl}[1]{\colorbox{YellowGreen}{\textbf{#1}}}

\title{MUtE: A Dual Framework for Concept Erasure \\ and Counterfactual Interventions}

\author{\name Antoine Saillenfest \email a.saillenfest@groupeonepoint.com \\
      \addr onepoint, 29 rue des Sablons, 75116 Paris (France)}

\def\month{MM}  
\def\year{YYYY} 
\def\openreview{\url{https://openreview.net/forum?id=XXXX}} 

\begin{document}

\maketitle

\begin{abstract}
 Erasing concept-specific information from representations has been proven useful for mitigating bias or interpreting model decisions. The joint objective is to transform the original representations such that the target concept becomes unpredictable, while maximally preserving concept-unrelated information. In this work, we revisit the optimal bounds of concept erasure to derive a novel class of erasure functions that naturally induce a deterministic, dual counterfactual mapping. Bridging the gap between theoretical optimality and practical representation learning, we design an implementation that imposes a translational bias on counterfactual trajectories—a constraint that aligns with how many concepts geometrically manifest in modern language models. Our framework enables seamless navigation between concept erasure and counterfactual generation. We empirically demonstrate its efficacy in improving downstream algorithmic fairness and generating counterfactual texts.
 \footnote{Code and data: \url{https://github.com/toinesayan/MUtE}.}
\end{abstract}

\section{Introduction}

Text representations inherently entangle a diverse array of latent concepts, ranging from sensitive demographic attributes (e.g. gender or race) requiring algorithmic mitigation to abstract semantic properties (e.g. writing style or aspect) manipulated for interpretability. Because targeted interventions in the discrete textual space are difficult to automate and frequently yield disfluent artifacts, representation engineering predominantly isolates and manipulates these concepts directly within the continuous embedding space.

The specific task of concept erasure aims to obliterate information pertaining to a target concept from a set of representations via an erasure function \citep{ravfogel2020null, ravfogel2022linear, chowdhury2025fundamental}. From an information-theoretic perspective, optimal erasure necessitates navigating a fundamental trade-off: maximizing privacy (ensuring the post-erasure representations contain minimal information regarding the target concept) while maximizing utility (maximizing the retained information from the original representations) \citep{chowdhury2025fundamental}. Because concept erasure is fundamentally task-agnostic, the resulting representations are broadly applicable across downstream NLP tasks, demonstrating efficacy in improving algorithmic fairness \citep{ravfogel2020null, chowdhury2022learning, lemberger2024explaining}, enhancing interpretability \citep{lemberger2024explaining}, and mitigating language model toxicity \citep{singh2024representation}.

\begin{figure}
    \centering
    \includegraphics[width=0.8\linewidth]{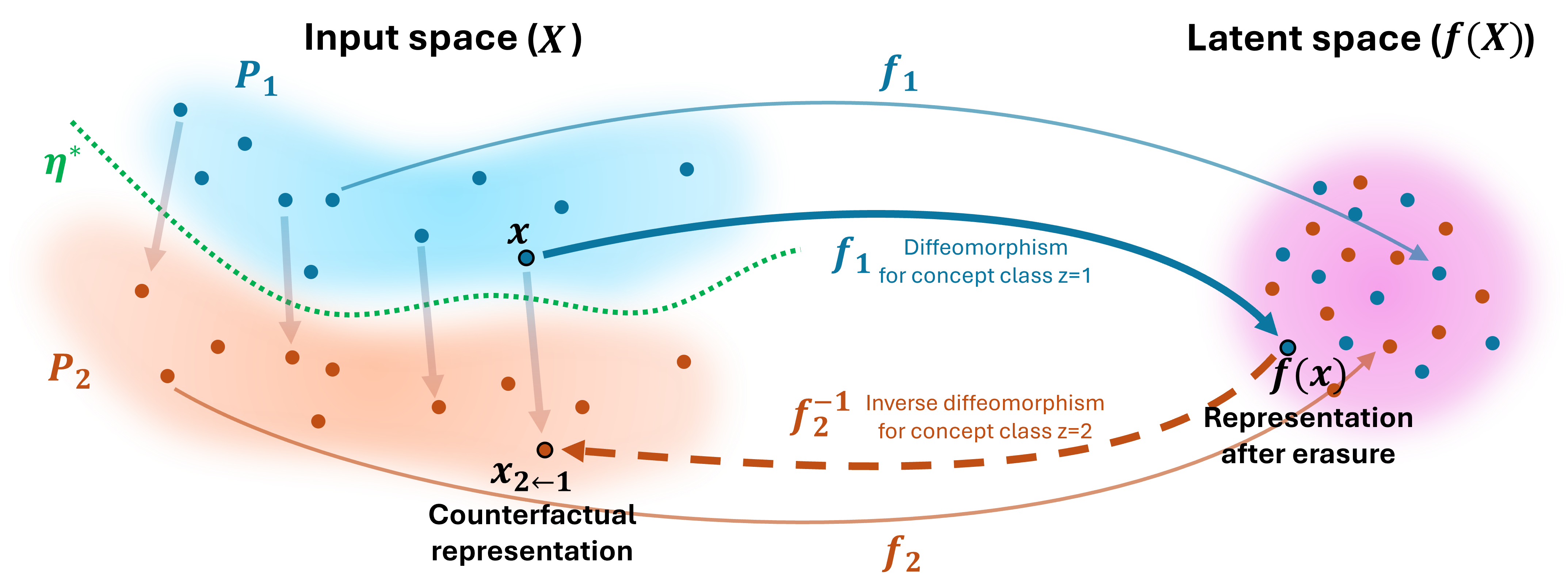}
    \caption{The $\text{MUtE}^*$ continuous erasure framework. A piecewise erasure function $f$ utilizes class-conditional diffeomorphisms $f_i$ to bijectively map source distributions $P_i$ to a shared invariant space, maximizing utility retention. This structure natively induces a dual counterfactual mapping ($f_2^{-1} \circ f_1$), anchoring a sample $x$ and its counterfactual $x_{2 \leftarrow 1}$ to the identical post-erasure representation $f(x)$. Constraining these counterfactual trajectories with an \textit{a priori} structural prior—such as the linear translational bias prevalent in NLP—ties the conditional mappings, effectively resolving the functional non-identifiability of optimal erasure.}
    \label{fig:motivation_img}
\end{figure}

Driven by these applications, concept erasure has seen rapid development. By relaxing strict privacy constraints, a significant body of work has addressed linear guardedness \citep{ravfogel2023log}, which consists in preventing concept recovery by a linear adversary. Methods such as INLP \citep{ravfogel2020null}, RLACE \citep{ravfogel2022linear}, and the state-of-the-art LEACE \citep{belrose2023leace} achieve linear erasure via projections onto a low-dimensional concept nullspace. However, general (non-linear) concept erasure remains an open challenge. Current approaches include adversarial training \citep{ganin2016domain, feder2021causalm, ravfogel2022adversarial}, information-theoretic loss optimization (FaRM \citep{chowdhury2022learning}, KRaM \citep{chowdhury2024robust}), Bayesian optimization (PEF \citep{chowdhury2025fundamental}), filtering of ranked directions out of embeddings (TaCo \cite{jourdan2023taco}) and density matching via projections (\leopard \citep{saillenfest2025nonlinear}). Those approaches, however, frequently sacrifice downstream utility to satisfy strict privacy bounds. Furthermore, these methods suffer from severe computational bottlenecks, relying on extensive neural network training, expensive density estimations, or Bayesian optimization over finite supports, which scales poorly in high-dimensional regimes.

Concurrently, recent literature has established profound connections between concept erasure and the generation of counterfactual representations, i.e. approximations of the embedding space under simulated causal interventions on the surface text. Assuming a structural causal model and Gaussian priors, \citet{lemberger2024explaining} have demonstrated that concept erasure can act as an intermediary mapping for counterfactual generation. Inspired by linear erasure techniques, recent works derive linear steering interventions under minimal displacement constraints \citep{singh2024representation}, which have been decoded into counterfactual texts using continuous-to-discrete inversion techniques \citep{avitan2025practical, morris2023text}.

In this work, we revisit discrete concept erasure within continuous representation spaces at optimality. We formalize the structure of optimal erasure functions and demonstrate that they inherently induce a dual, deterministic counterfactual mapping. Bridging theory and practice, we propose a practical implementation that imposes a rigid translational bias on counterfactual trajectories. This approach practically exploits the phenomenon that modern text encoders frequently isolate concepts as linear directions.

To summarize, our primary contributions are:
\begin{itemize}
    \item We introduce \textbf{MUtE$^*$} (\textbf{M}aximum \textbf{Ut}ility-preserving \textbf{E}rasure), a class of optimal erasure functions for continuous representations that intrinsically defines a dual counterfactual mapping (Section \ref{sec:erasure_and_counterfactuals}).
    \item We derive a practical, computationally efficient implementation that leverages a translational bias to perform erasure and counterfactual representation generation (Section \ref{sec:practical_implementation}).
    \item We empirically validate our framework across synthetic data and NLP benchmarks, demonstrating its efficacy for bias mitigation and counterfactual text generation (Section \ref{sec:experiments}).
\end{itemize}

\section{Optimal Erasure and Counterfactuals}
\label{sec:erasure_and_counterfactuals}

This section formalizes the theoretical foundations of Maximum Utility-preserving Erasure ($\text{MUtE}^*$). After defining the problem setting (Section \ref{sec:problem_statement}), we review the necessary conditions for perfect erasure (Section \ref{sec:discrete_concept_erasure}) and maximal utility retention (Section \ref{sec:maximum_utility_preservation}). Assuming an oracle predictor, we demonstrate that this optimal erasure naturally induces a dual counterfactual mapping (Section \ref{sec:counterfactual_mapping}), providing a principled geometric mechanism to resolve the inherent non-identifiability of erasure functions.

\subsection{Problem statement}
\label{sec:problem_statement}

Let $\mathbf{P}_{\mathcal{S}}$ be a probability distribution over an arbitrary input space $\mathcal{S}$. Let $\mathrm{enc}: \mathcal{S} \to \mathbb{R}^D$ be an encoding function mapping inputs to a continuous representation space, and let $\mathcal{Z} = \{1, \dots, K \}$ ($K > 1$) denote a discrete set of concept classes assigned by a ground-truth labeling function $c^*: \mathcal{S} \to \mathcal{Z}$.

We define the concept variable $Z$ and the representation variable $X \sim \mathbf{P}_X$ as random variables over $\mathcal{S}$:
\begin{equation}
    \label{eq:Z}
    Z(s) = c^*(s)
\end{equation}
\begin{equation}
    \label{eq:X}
    X(s) = \mathrm{enc}(s)
\end{equation}

Let $p(x)$ denote the probability density function (PDF) of $\mathbf{P}_X$. For each concept class $i \in \mathcal{Z}$, we introduce the class-conditional representation variable $X_i \sim \mathbf{P}_i$, where the corresponding conditional density is defined as $p_i(x) := p(x \mid Z = i)$.

Concept erasure seeks an \textbf{erasure function} $f$ that removes concept-related information about $Z$ (Eq. \ref{eq:Z}) from the representations $X$ (Eq. \ref{eq:X}). We define the post-erasure representation as $Q = f(X)$, distributed according to the pushforward measure $\mathbf{Q} = f_{\#} \mathbf{P}_X$, with corresponding class-conditional distributions $\mathbf{Q}_i$.

We make the following assumptions:
\begin{itemize}
    \item The erasure function preserves the latent dimensionality, i.e., $f: \mathbb{R}^D \to \mathbb{R}^D$.
    \item The conditional distributions $\mathbf{P}_i$ are continuous over $\mathbb{R}^D$ and exhibit full support (i.e., there is no zero-density regions).\footnote{In our primary NLP setting ($\mathcal{S} = \Sigma^*$, where $\Sigma$ is an alphabet and $\mathbf{P}_{\mathcal{S}}$ is a language model), discrete textual inputs strictly map to point masses in $\mathbb{R}^D$. To satisfy continuity and full-support constraints, we assume these discrete representations are continuously smoothed (e.g., via kernel density estimation).}
\end{itemize}

\subsection{Discrete Concept Erasure.} 
\label{sec:discrete_concept_erasure}

Concept erasure fundamentally seeks a transformation $f$ that enforces strict statistical independence between $f(X)$ and $Z$. This translates into a rigorous distributional constraint: the class-conditional distributions of the post-erasure representations must be perfectly aligned \citep{saillenfest2025nonlinear}. Formally:

\begin{equation}
\label{eq:concept_erasure}
    \forall i\in \mathcal{Z}, \quad \mathbf{Q}_i =  \mathbf{Q}
\end{equation}

 \noindent From an information-theoretic perspective, Eq. \ref{eq:concept_erasure} corresponds to achieving the privacy bound characterized by vanishing mutual information: $I(f(X); Z) = 0$ \citep{chowdhury2025fundamental}.
 
 Crucially, however, this pure erasure objective is not self-sufficient. An infinite set of mappings satisfies this independence criterion, including degenerate, constant functions (e.g., $f(x) = k \in \mathbb{R}^D$) that catastrophically collapse the representation space. Consequently, any viable concept erasure framework must formulate a joint objective: satisfying the erasure criterion while maximizing the retention of task-agnostic information (utility) inherent to the original representations.

\subsection{Maximum Utility Preservation.}
\label{sec:maximum_utility_preservation}

From an information-theoretic perspective, preserving maximum utility requires the conditional entropy to vanish: $H(X|f(X),Z) = 0$ \citep{chowdhury2025fundamental}. Conditioning on a concept class $i \in \mathcal{Z}$, the equality $H(X_i|f(X_i)) = 0$ implies that $X_i$ can be perfectly recovered from its post-erasure representation. Since the forward mapping is deterministic by definition ($H(f(X_i)|X_i) = 0$), this mutual determinism ensures that $f$ maps each conditional distribution $\mathbf{P}_i$ bijectively to $\mathbf{Q}_i$.

Structurally, integrating the perfect erasure constraint (Eq. \ref{eq:concept_erasure}), \textbf{MUtE$^*$} (Maximum Utility-preserving Erasure) functions satisfies:

\begin{equation}
    \label{eq:MUtE_s_dependant}
    \forall s \in \mathcal{S}, \quad f(X(s)) = f_{c^*(s)}(X(s)) 
\end{equation}
\noindent where each $f_i : \mathbb{R}^D \to \mathbb{R}^D$ is a diffeomorphism transforming $\mathbf{P}_i$ into $\mathbf{Q}$.

This continuous-space formulation mirrors the optimal erasure framework introduced by \citet{chowdhury2025fundamental}. Their approach achieves optimal privacy-utility bounds via class-conditional permutations, assuming discrete supports of equal cardinality that are identical up to permutation relative to the target distribution. Our framework generalizes these optimality principles to continuous probability measures by replacing combinatorial permutations with diffeomorphisms.

The sample-dependent assignment $c^*(s)$ in Eq. \ref{eq:MUtE_s_dependant} precludes $f$ from operating solely on the observable representation $X$. To circumvent this limitation, \citet{chowdhury2025fundamental} assume disjoint conditional supports, guaranteeing perfect predictability of the concept class from the representation alone. In practice, high-dimensional continuous representations inevitably exhibit overlapping supports, yielding an irreducible Bayes error that precludes strict determinism. However, because modern latent spaces demonstrate high structural separability, routinely allowing probes to achieve near-perfect empirical classification, we adopt a tractable theoretical surrogate. We assume the existence of an optimal oracle predictor $\eta^*$ that recovers the concept class with negligible error, idealizing it as exact for our formulation:
\begin{equation}
    \label{eq:oracle}
    \forall s \in \mathcal{S}, \quad \eta^*(X(s)) = Z(s) = c^*(s)
\end{equation}

Under this oracle assumption, $\text{MUtE}^*$ functions (Eq. \ref{eq:MUtE_s_dependant}) can be expressed as a class of representation-dependent erasure mappings:
\begin{equation}
    \label{eq:MUtE}
    f : \mathbb{R}^D \to \mathbb{R}^D, \quad x \mapsto f_{\eta^*(x)}(x)
\end{equation}

Because concept erasure enforces alignment strictly at the distributional level (Eq. \ref{eq:concept_erasure}), the hypothesis space of valid $\text{MUtE}^*$ mappings is infinite. This permits arbitrary topological distortions of the representation space. To resolve this fundamental non-identifiability, we next exploit the dual counterfactual mapping intrinsically induced by these optimal erasure functions.

\subsection{Dual Counterfactual Mapping}
\label{sec:counterfactual_mapping}

Let the counterfactual mapping of a representation-dependent MUtE$^*$ function $f$ (Eq. \ref{eq:MUtE}) be:
\begin{equation}
\label{eq:counterfactual_mapping}
f_{\leftarrow} : \mathbb{R}^D \times \mathcal{Z} \to \mathbb{R}^D, \quad (x, j) \mapsto f_j^{-1}(f(x))
\end{equation}

This transport is termed \textit{counterfactual} because, for any distinct concept classes $i, j \in \mathcal{Z}$, it maps $X_i \sim \mathbf{P}_i$ to $X_{j\leftarrow i} := f_\leftarrow(X_i, j) = f_j^{-1}(f_i(X_i))$ that is strictly distributed according to $\mathbf{P}_j$. 

By definition, $f(X_{j\leftarrow i}) = f(X_i)$. This establishes a fundamental structural correspondence between a MUtE$^*$ function and the counterfactuals it generates: a sample and all its counterfactual counterparts map to the exact same invariant coordinate in the post-erasure latent space. Consequently, each erased representation serves as a geometric anchor connecting $|\mathcal{Z}|$ counterfactual representations.

Crucially, this collision property provides a principled mechanism to resolve the inherent non-identifiability of optimal erasure functions. If the geometry of the target counterfactuals $X_{j\leftarrow i}$ can be approximated \textit{a priori}, it directly constrains the hypothesis space of $f$ by tying the diffeomorphisms $f_i$ and $f_j$. By enforcing $f(X_{j\leftarrow i}) = f(X_i)$ over these expected counterfactual trajectories, we can isolate geometrically grounded solutions from the infinite space of otherwise valid MUtE$^*$ functions.

In NLP, the geometric relationship between representations and their counterfactuals has been extensively studied. The Linear Bias Hypothesis posits that deep NLP models naturally encode concepts as linear directions within their high-dimensional latent spaces \citep{bolukbasi2016man, vargas2020exploring}. Thus, recent steering techniques operationalize counterfactual generation via linear interventions \citep{singh2024representation}. The simplest such intervention approximates a counterfactual shift by translating the representation along the vector difference between class centroids \citep{subramani2022extracting, singh2024representation}.

Synthesizing the theoretical formulation above, Figure \ref{fig:motivation_img} illustrates the $\text{MUtE}^*$ framework, demonstrating how its invertible mappings enable bidirectional geometric navigation between concept erasure and counterfactual generation.

\section{Operationalizing \texorpdfstring{MUtE$^*$}{MUtE*} via Iterative Density Matching}
\label{sec:practical_implementation}

This section bridges the theoretical formulation of $\text{MUtE}^*$ with a computationally tractable implementation. We first recall an iterative Gaussianization procedure (Section \ref{sec:RBIG}), adapting it to construct a fully representation-dependent erasure mapping (Section \ref{sec:MUtE_empirical}). Geometrically, this mapping imposes an inherent translational bias on the counterfactual trajectories. We then relax the assumption of an oracle predictor $\eta^*$ to formally characterize the impact of noisy routing. This theoretical analysis directly motivates algorithmic regularizations to promote effective concept erasure within deeply entangled latent spaces (Section \ref{sec:algorithmic_mitigation}).

\subsection{Iterative Gaussianization}
\label{sec:RBIG}

Rotation-Based Iterative Gaussianization (RBIG)  \citep{laparra2011iterative} is a highly efficient, iterative Gaussianization technique that transforms any continuous random distribution $\mathbf{P}$ into an isotropic Gaussian $\mathcal{N}(0, I)$. For a set of observations $x$, the process alternates between applying an orthogonal rotation matrix $R^{(t)} \in \mathrm{O}(D)$ and a dimension-wise marginal Gaussianization $\psi^{(t)}$:

\begin{equation}
\label{eq:RBIG}
\begin{array}{l}
     x^{(t+1)} = \left(\psi^{(t)} \circ R^{(t)}\right)(x^{(t)}) \\ \text{with}\quad x^{(0)} = x      
\end{array}
\end{equation}

\noindent where:

\begin{equation}
    \label{eq:marginal_gaussianization}
    \psi^{(t)}(x^{(t)}) = \left( \Phi^{-1} \left(\int_{-\infty}^{x^{(t)}_d} p^{(t)}_d(u) du \right) \right)_{d=1,\dots,D}
\end{equation}

The marginal Gaussianization $\psi^{(t)}$ operates independently on each dimension $d$, mapping the data to $\mathcal{N}(0, 1)$ via a marginal uniformization, based on the cumulative density function (CDF) of the marginal probability density function (PDF) $p_d$, followed by the inverse CDF of the standard normal $\mathcal{N}(0,1)$, $\Phi^{-1}$.

Provided the sequence of orthogonal rotations $\{R^{(t)}\}$ induces sufficient cross-dimensional mixing (e.g., via orthogonal ICA \citep{hyvarinen2019nonlinear}, PCA \citep{jolliffe2016principal}, or random orthogonal matrices), RBIG guarantees monotonic convergence to $\mathcal{N}(0, I)$. This monotonic convergence is formally characterized by a strict reduction in negentropy, defined here as the Kullback-Leibler divergence to the standard isotropic Gaussian, at step $t$. The negentropy reduction $\Delta J^{(t)}$ at step $t$ is thus strictly positive:


\begin{equation}
    \label{eq:negentropy_reduction}
    \Delta J^{(t)} := D_{\mathrm{KL}}(\mathbf{P}^{(t)} \parallel \mathcal{N}(0, I)) - D_{\mathrm{KL}}(\mathbf{P}^{(t+1)} \parallel \mathcal{N}(0, I)) > 0
\end{equation} 

The overall Gaussianization process is bijective. Rotations are invertible and $(R^{(t)})^{-1}= (R^{(t)})^\top$. $\psi^{(t)}$ is invertible when the support of each marginal PDF is connected (i.e. there are no zero-probability regions) making the marginal CDF strictly monotonic and hence invertible.

\subsection{Selective Iterative Density Matching}
\label{sec:MUtE_empirical}

We adapt the process defined in Eq. \ref{eq:RBIG} into a conditional mapping, selective based on the predicted class $\eta^*(x)$ of the sample $x$:

\begin{equation}
\label{eq:SIG}
\begin{array}{l}
     x^{(t+1)} = \left(\psi_{\eta^*(x)}^{(t)} \circ R^{(t)}\right)(x^{(t)}) \\ \text{with}\quad x^{(0)} = x 
\end{array}
\end{equation}

\noindent where $R^{(t)}$ is an orthogonal rotation, and $\psi_{i}^{(t)}$ denotes the class-conditional, dimension-wise marginal Gaussianization at step $t$. For a given concept class $i$, $\psi_{i}^{(t)}$ is defined using the marginal class-conditional PDF $p_{i,d}^{(t)}$ for the dimension $d$:

\begin{equation}
    \label{eq:conditional_marginal_gaussianization}
    \psi_{i}^{(t)}(x) = \left( \Phi^{-1} \left( \int_{-\infty}^{x_d} p_{i,d}^{(t)}(u) du \right) \right)_{d=1,\dots,D} 
\end{equation}

At each iteration, $R^{(t)}$ is applied uniformly across the entire representation space, acting as an isometry that preserves the macroscopic geometric structure. Conversely, the class-conditional marginal transformations $\psi_{i}^{(t)}$ independently warp the conditional densities to align their marginals with a standard normal distribution.

Under the assumption that $\eta^*$ acts as a perfect oracle (Eq. \ref{eq:oracle}) and conditioned on any concept class $i$, this process mirrors the standard RBIG, which guarantees that $\mathbf{P}_i$ is bijectively mapped to an isotropic Gaussian $\mathcal{N}(0,I)$ as $t \to \infty$. Consequently, this iterative procedure asymptotically drives all conditional distributions $\mathbf{P}_i$ to a shared target measure, theoretically guaranteeing complete concept erasure after a sufficient number of iterations $T$.\footnote{Concept erasure requires only that the class-conditional densities match (Eq. \ref{eq:concept_erasure}), not full convergence to $\mathcal{N}(0,I)$. In practice, $T$ denotes the number of steps to satisfy Eq. \ref{eq:concept_erasure}.}

Truncating the iterative process at step $T$ yields an empirical conditional-bijective \textbf{MUtE} function:

\begin{equation}
    \label{eq:theoretical_MUtE}
    \begin{array}{l}
    f:\R^{d} \to \R^{d}, \quad x\mapsto f_{\eta^*(x)}(x)\\
    \text{with }f_{i}=\psi_{i}^{(T-1)} \circ R^{(T-1)} \circ \dots \circ \psi_{i}^{(0)} \circ  R^{(0)} 
    \end{array}
\end{equation}

Crucially, the initial marginal Gaussianizations $(\psi_{i}^{(0)})_{i=1, \dots, |\mathcal{Z}|}$ independently map the median of each conditional distribution to the origin. Because deep continuous representations typically exhibit symmetric, Gaussian-like marginals where the mean and median closely align, this zero-order matching acts as a rigid location shift. Consequently, this dominant first step imparts a strong geometric inductive bias, steering counterfactual representations predominantly along the inter-centroid vector $\mathbb{E}[X_j] - \mathbb{E}[X_i]$. The subsequent sequence of global rotations and marginal transformations non-linearly refines this trajectory to ensure exact higher-order distributional alignment without destroying the foundational translational bias.

\subsection{Implementation and Mitigation of Noisy Routing}
\label{sec:algorithmic_mitigation}

Due to overlapping conditional supports, the empirical predictor $\eta^*$ inherently exhibits classification error. Crucially, $\text{MUtE}$ decouples density estimation from sample routing: marginal transformations $\psi_{i}^{(t)}$ are fitted using ground-truth labels, while the forward mapping of a sample $x$ relies strictly on its predicted class $\eta^*(x)$.

This irreducible Bayes error theoretically precludes strict convergence to an exact isotropic Gaussian. We formalize this limitation by expressing the stepwise negentropy reduction for class $i$ under noisy routing (see Appendix \ref{sec:app:noisy_routing} for the full derivation):

\begin{equation}
\label{eq:negentropy_evolution_noisy_routing}
\Delta J_i^{(t)} = \Delta J^{*(t)}_i - E^{(t)}_i
\end{equation}

where $\Delta J^{*(t)}_i \geq 0$ denotes the ideal negentropy reduction at step $t$ under perfect routing, and $E^{(t)}_i > 0$ represents a strictly positive entropic penalty induced by misrouting. Consequently, the convergence of each class-conditional distribution to $\mathcal{N}(0, I)$ is fundamentally bottlenecked by this routing error.

Nevertheless, this analytical decomposition directly motivates algorithmic regularizations to optimize the stepwise negentropy reduction—specifically, by maximizing the ideal marginal gain $\Delta J^{*(t)}_i$ and bounding the routing penalty $E^{(t)}_i$. Although exact asymptotic convergence to the standard normal is mathematically unattainable, we expect that enforcing these bounds iteratively drives the conditional manifolds into a sufficiently tight, shared neighborhood, to effectively neutralizes class separability.

To maximize the marginal Gaussianization gain $\Delta J^{*(t)}_i$, the rotation sequence $\{R^{(t)}\}$ must systematically induce sufficient cross-dimensional mixing. While techniques such as orthogonal ICA drive rapid convergence, PCA offers a superior trade-off between step-wise negentropy reduction and computational complexity \citep{laparra2011iterative}. To satisfy this mixing requirement concurrently across all concept classes $i \in \mathcal{Z}$, we adopt an alternating rotation schedule: $R^{(t)} = R_{i}^{\text{PCA}}$, where $i \equiv t \pmod{|\mathcal{Z}|}$ and $R_{i}^{\text{PCA}}$ diagonalizes the covariance of $\mathbf{P}_i^{(t)}$. This cyclic scheme periodically aligns the principal axes of each conditional manifold, ensuring their unique cross-correlations are successively exposed for marginal Gaussianization. Although a rotation $R_{i}^{\text{PCA}}$ optimized for class $i$ does not explicitly target the cross-correlations of a distinct class $j \neq i$, the RBIG framework guarantees that any valid orthogonal rotation still yields a non-negative negentropy reduction ($\Delta J^{*(t)}_j \geq 0$). Moreover, if the class-conditional manifolds are approximately equivalent up to a translation, their covariance structures inherently align. In this regime, rotating by $R_{i}^{\text{PCA}}$ is thus expected to simultaneously drive a high negentropy reduction for all classes at every iterative step.

The routing penalty $E^{(t)}_i$ quantifies the entropic cost of spatial tearing, driven by the cross-entropy mismatch when a sample is evaluated via the marginal density estimator of a mispredicted class. To maintain focus on the practical mitigation strategy, we defer the full analytical expansion and bounding of $E^{(t)}_i$ to Appendix \ref{app:negentropy_reduction_noisy_routing}. Crucially, this penalty diverges to infinity if the true density is strictly positive where the mispredicted estimator assigns zero probability. To bound this log-density ratio, we estimate the continuous conditional PDFs $p_{i,d}^{(t)}$ via marginal histograms uniformly discretized into $B$ bins, enforcing a strict minimum density threshold $p_{i,d}^{(t)}(x_d) \geq \alpha > 0$. This threshold constrains $E^{(t)}_i$ with an upper bound. By defaulting to $\alpha = 10^{-10}$ and $B=1000$, the artificial probability mass injected across the domain ($B \alpha = 10^{-7}$) remains negligible.

While these regularizations cannot theoretically guarantee perfect distributional alignment after $T$ iterations, our subsequent evaluations on real-world datasets demonstrate that this constrained implementation yields highly robust concept erasure in practice.

\section{Experiments}
\label{sec:experiments}

We validate our approach on synthetic and NLP datasets (Section \ref{sec:datasets_baselines_training}) for concept erasure and utility preservation (Section \ref{sec:erasure_utility_preservation}), bias mitigation (Section \ref{sec:fair_classification}) and counterfactual generation of texts (Section \ref{sec:counterfactual_text_generation}). 

\subsection{Datasets, Baselines and Training details}
\label{sec:datasets_baselines_training}

\begin{table}[ht!]
    \centering
    \caption{Key dataset statistics. $y$ denotes the availability of a downstream classification task.}
    \begin{tabular}{llrlcrrr}
        \\
        \hline
                Dataset& Encoder & $D$ & Concept & $|\mathcal{Z}|$ & \#train &\#test & $y$\\
        \hline
        \textsc{GloVe} & GloVe & 300 & Gender & 3 & \textasciitilde 11k & \textasciitilde 7k &  \\
        \textsc{Bias in Bios} & Bert & 768 & Gender & 2 & \textasciitilde 256k & \textasciitilde 98k & \checkmark\\
        \textsc{DIAL} & Deepmoji & 300 & Race & 2 & \textasciitilde 180k & \textasciitilde 8k & \checkmark\\
        \textsc{Jigsaw} & GPT-4 & 512 & Religion & 5 & \textasciitilde 87k & \textasciitilde 9k & \checkmark\\
        \hline
    \end{tabular}
    \label{tab:datasets-short-details}
\end{table}

\paragraph{Baselines.} We compare our approach with several baselines for erasure: FaRM \citep{chowdhury2022learning}, KRaM \citep{chowdhury2024robust}, TaCo \citep{jourdan2023taco}, and $\overline{\mathrm{L}}$EOPARD \citep{saillenfest2025nonlinear}, and the linear erasure method LEACE \citep{belrose2023leace} for completeness. Training details are deferred to Appendix \ref{sec:app:training-parameters}.

\paragraph{NLP benchmarks.} We evaluate our approach for concept erasure across a suite of text embeddings: gender from \textsc{GloVe} \citep{pennington2014glove}, gender from BERT representations \citep{devlin2019BERT} of short biographies in \textsc{Bias in Bios} \citep{dearteaga2019bias}, race from DeepMoji representations of \textsc{DIAL} tweets \citep{blodgett2016demographic}, and religion from GPT-4 embeddings \citep{achiam2023gpt} of online comments in \textsc{Jigsaw} \citep{jigsaw2019}. Table \ref{tab:datasets-short-details} summarizes key statistics, other details are deferred to Appendix \ref{sec:app:datasets}.

\paragraph{Evaluation.} Probes for concept prediction and downstream classification were implemented as MLPs (\texttt{scikit-learn}'s \texttt{MLPClassifier} \citep{pedregosa2011scikit}). Mean Squared Error (MSE) estimations were conducted using \texttt{MLPRegressor}. Accuracies, MSEs, and fairness scores reported are averages across $5$ independent evaluations.

\begin{figure}[t!]
    \centering
    \includegraphics[width=\linewidth]{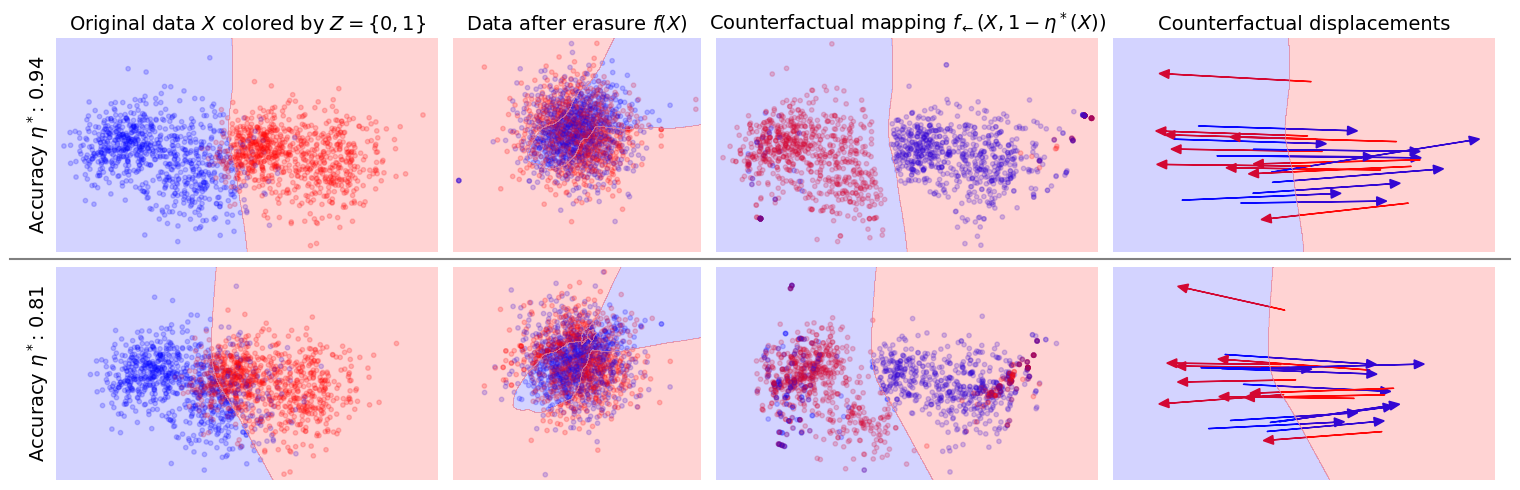}
    \caption{Concept erasure via $\text{MUtE}$ on synthetic distributions exhibiting near-perfect separability (top row) versus moderate overlap (bottom row). Background hues indicate decision regions for the routing predictor (cols. 1, 3, 4) and the optimal adversarial probe (col. 2). Point colors denote ground-truth concept labels.}
    \label{fig:synthetic}
\end{figure}

\begin{figure}[t!]
    \centering
    \includegraphics[width=0.8\columnwidth]{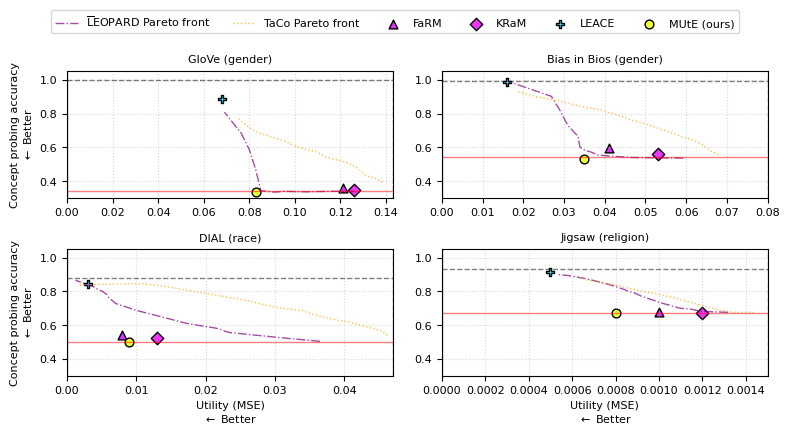}
    \caption{Probing accuracy-Utility tradeoff. Horizontal red (resp. dotted) lines are chance-level baseline accuracies (resp. accuracies on the original latent space).}
    \label{fig:utility_privacy_tradeoff}
\end{figure}

\subsection{Erasure and utility preservation}
\label{sec:erasure_utility_preservation}

\paragraph{Synthetic data.}
We evaluate MUtE on synthetic $\mathbb{R}^2$ data, modeling a binary concept using two different Gaussian mixtures of $2,000$ samples each. We control concept separability by translating the mixtures along a fixed vector, defining two regimes with initial probe accuracies of $94\%$ and $81\%$, respectively (Figure \ref{fig:synthetic}, left). In both settings, we train MUtE for $20$ iterations. Post-erasure, the conditional distributions are aligned, and probe accuracy drops to chance (Figure \ref{fig:synthetic}, second column). In both regimes, the mapping produces artifacts corresponding to misclassified samples; their counterfactual displacement is roughly inverse to the trajectory expected under ground-truth routing (Figure \ref{fig:synthetic}, third column). The counterfactual mapping exhibits a strong translational bias, confirming our geometric priors (Figure \ref{fig:synthetic}, right).

\paragraph{Real-world datasets}

We extend our evaluation of MUtE to real-world NLP benchmarks. Results in Figure \ref{fig:utility_privacy_tradeoff} indicates that MUtE consistently reduces probe accuracy for sensitive attributes to the majority-class baseline (e.g., $50\%$ for balanced binary concepts) consistently outperforming baseline methods. Following \citet{chowdhury2025fundamental}, we use the MSE of reconstructing $X$ from $f(X)$ as a surrogate for utility preservation, noting that MUtE maintains low reconstruction error. Furthermore, MUtE demonstrates competitve or higher performance (accuracy $a_y$) on downstream tasks (Table \ref{tab:fairness_results}).

For \textsc{GloVe}, we quantify the preservation of intrinsic semantic and local topological properties. On the WordSim-353 benchmark, we measure the Spearman rank correlation between embedding cosine similarities and human annotations \citep{agirre2009study}. The original embeddings yield a correlation of $0.70$. MUtE successfully preserves this semantic alignment ($0.71$), whereas baseline methods degrade it ($\text{FaRM}: 0.53$, $\text{KRaM}: 0.33$). We also evaluate local topological fidelity by computing the retention rate of the top $1\%$ nearest neighbors post-erasure. MUtE retains $19\%$ of the original neighborhood structure, outperforming both $\text{FaRM}$ ($13\%$) and $\text{KRaM}$ ($10\%$).

\begin{table}[t!]
    \centering
    \caption{Fairness evaluation for downstream classifiers. Best results (excluding original representations) in bold.\\}
    \begin{tabular}{lrrr}
        \toprule
        \textbf{Model} &   \bm{$a_y$} (\%) $\uparrow$ &  \bm{$\mathrm{TPR}^{\mathrm{RMS}}$} $\downarrow$ & DP $\downarrow$ \\ 
        
        \midrule
        \multicolumn{4}{l}{\textsc{Bias in Bios}} \\
        \textit{orig.} & \textit{80.0{\scriptsize $\;\pm\;$0.1}} & \textit{0.164{\scriptsize $\;\pm\;$0.009}} & \textit{0.572{\scriptsize $\;\pm\;$0.005}} \\
        FaRM & 55.4{\scriptsize $\;\pm\;$0.1} & 0.082{\scriptsize $\;\pm\;$0.004} & 0.314{\scriptsize $\;\pm\;$0.002} \\
        KRaM & 51.2{\scriptsize $\;\pm\;$0.3} & \textbf{0.056}{\scriptsize $\;\pm\;$\textbf{0.003}} & \textbf{0.285}{\scriptsize $\;\pm\;$\textbf{0.004}} \\
        \rowcolor{black!5}MUtE & \textbf{70.2}{\scriptsize $\;\pm\;$\textbf{0.2}} & 0.091{\scriptsize $\;\pm\;$0.002} & 0.391{\scriptsize $\;\pm\;$0.004} \\

        \midrule
        \multicolumn{4}{l}{\textsc{DIAL}} \\
        \textit{orig.} & \textit{75.8{\scriptsize $\;\pm\;$0.1}} & \textit{0.156{\scriptsize $\;\pm\;$0.007}} & \textit{0.269{\scriptsize $\;\pm\;$0.017}} \\
        FaRM & 73.3{\scriptsize $\;\pm\;$0.5} & \textbf{0.079}{\scriptsize $\;\pm\;$\textbf{0.004}} & 0.061{\scriptsize $\;\pm\;$0.005} \\
        KRaM & \textbf{73.1}{\scriptsize $\;\pm\;$\textbf{0.2}} & \textbf{0.084}{\scriptsize $\;\pm\;$\textbf{0.005}} & \textbf{0.006}{\scriptsize $\;\pm\;$\textbf{0.006}} \\
        \rowcolor{black!5}MUtE & 71.0{\scriptsize $\;\pm\;$0.3} & 0.094{\scriptsize $\;\pm\;$0.005} & 0.086{\scriptsize $\;\pm\;$0.012} \\
        \bottomrule
    \end{tabular}
    \label{tab:fairness_results}
\end{table}

\subsection{Fair classification}
\label{sec:fair_classification}

A primary downstream application of concept erasure is mitigating algorithmic bias to improve fairness. Following established methodologies \citep{ravfogel2020null, dearteaga2019bias, saillenfest2025nonlinear}, we evaluate fairness using two metrics suited for binary concepts: the Root Mean Square of the True Positive Rate Gap ($\mathrm{TPR}^{\mathrm{RMS}}$) and Demographic Parity (DP) (those metrics are formally defined in Appendix \ref{sec:app:fairness_metrics}). Results in Table \ref{tab:fairness_results} shows that downstream classifiers trained on MUtE-transformed representations exhibit substantial fairness improvements compared to those trained on the original latent space. MUtE is competitive with established baselines, establishing a distinct operating point on the accuracy-fairness Pareto frontier that strongly favors the preservation of representation utility for \textsc{Bias in Bios}.

The dual counterfactual mapping induced by MUtE enable its application as a data augmentation technique to train fair classifiers directly within the original representation space. Specifically, we train downstream classifiers on \textsc{Bias in Bios} and \textsc{DIAL} using 50k original samples paired with their 50k generated counterfactuals, yielding a balanced 100k-sample training corpus. We evaluate two distinct supervisory regimes: (1) Oracle-Supervised (MUtE$^*_\leftarrow$): Concept and task labels are jointly available during training, allowing us to generate exact counterfactual representations using the theoretical MUtE$^*$ function, and (2) Disjoint-Supervised (MUtE$_\leftarrow$): Concept and task labels reside in mutually exclusive training sets, MUtE is optimized strictly on the concept-annotated partition and subsequently deployed to augment the task-annotated partition. We benchmark these approaches against two latent steering interventions for counterfactual generation: class-conditional centroid translation (Mean Diff.) \citep{subramani2022extracting, singh2024representation} and class-conditional linear optimal transport (Linear OT) \citep{singh2024representation}. As shown in Table \ref{tab:data_augmentation_fairness_results}, counterfactual augmentation via MUtE$^*_\leftarrow$ and MUtE$_\leftarrow$ drives significant fairness gains in the original representation space, consistently rivaling or outperforming linear steering techniques.

\begin{table}[t]
    \centering
    \caption{Fairness evaluation after data augmentation. Best results (excluding original representations) in bold.\\}
    \begin{tabular}{lrrr}
        \toprule
        \textbf{Model} &   \bm{$a_y$} (\%) $\uparrow$ &  \bm{$\mathrm{TPR}^{\mathrm{RMS}}$} $\downarrow$ & DP $\downarrow$ \\ 
        
        \midrule
        \multicolumn{4}{l}{\textsc{Bias in Bios}} \\
        \textit{orig.} & \textit{78.3{\scriptsize $\;\pm\;$0.2}} & \textit{0.171{\scriptsize $\;\pm\;$0.010}} & \textit{0.568{\scriptsize $\;\pm\;$0.007}} \\
        Mean diff. & 73.9{\scriptsize $\;\pm\;$0.2} & 0.136{\scriptsize $\;\pm\;$0.007} & 0.534{\scriptsize $\;\pm\;$0.005} \\
        Linear OT & 73.3{\scriptsize $\;\pm\;$0.2} & \textbf{0.099}{\scriptsize $\;\pm\;$\textbf{0.005}} & \textbf{0.449}{\scriptsize $\;\pm\;$\textbf{0.006}} \\
        \rowcolor{black!5}MUtE$^*_\leftarrow$ & \textbf{75.6}{\scriptsize $\;\pm\;$\textbf{0.6}} & \textbf{0.099}{\scriptsize $\;\pm\;$\textbf{0.003}} & \textbf{0.464{\scriptsize $\;\pm\;$0.010}} \\
        \rowcolor{black!5}MUtE$_\leftarrow$ & \textbf{76.1}{\scriptsize $\;\pm\;$\textbf{0.4}} & \textbf{0.105}{\scriptsize $\;\pm\;$\textbf{0.006}} & 0.470{\scriptsize $\;\pm\;$0.013} \\

        \midrule
        \multicolumn{4}{l}{\textsc{DIAL}} \\
        \textit{orig.} & \textit{75.8{\scriptsize $\;\pm\;$0.1}} & \textit{0.155{\scriptsize $\;\pm\;$0.003}} & \textit{0.268{\scriptsize $\;\pm\;$0.006}} \\
        Mean diff. & \textbf{75.5{\scriptsize $\;\pm\;$0.1}} & 0.154{\scriptsize $\;\pm\;$0.005} & 0.265{\scriptsize $\;\pm\;$0.010} \\
        Linear OT & \textbf{75.5}{\scriptsize $\;\pm\;$\textbf{0.2}} & 0.144{\scriptsize $\;\pm\;$0.010} & 0.243{\scriptsize $\;\pm\;$0.024} \\
        \rowcolor{black!5}MUtE$^*_\leftarrow$ & \textbf{75.3{\scriptsize $\;\pm\;$0.3}} & \textbf{0.114{\scriptsize $\;\pm\;$0.006}} & \textbf{0.168{\scriptsize $\;\pm\;$0.014}} \\
        \rowcolor{black!5}MUtE$_\leftarrow$ & \textbf{75.6{\scriptsize $\;\pm\;$0.1}} & \textbf{0.109}{\scriptsize $\;\pm\;$\textbf{0.007}} & \textbf{0.154}{\scriptsize $\;\pm\;$\textbf{0.017}} \\

        \bottomrule
    \end{tabular}
    \label{tab:data_augmentation_fairness_results}
\end{table}

\subsection{Counterfactual text generation}
\label{sec:counterfactual_text_generation}

We evaluate the quality of counterfactual representations generated by $\text{MUtE}^*_\leftarrow$ against linear steering baselines in the context of discrete text generation. Using the continuous-to-discrete text inversion framework proposed by \citet{morris2023text}, we project the counterfactual embeddings back into the natural language space. Specifically, we train a 4-iteration $\text{MUtE}^*$ model on GTR embeddings \citep{ni2022large} derived from the \textsc{Bias in Bios} dataset. For test sequences under $32$ tokens ($N=628$, bounded by the maximum sequence length observed during the inversion model's training), we generate counterfactual embeddings using $\text{MUtE}^*_\leftarrow$, Mean Difference, and Linear OT, which are subsequently decoded into text. To isolate the impact of the optimization objective from the functional capacity of the mapping, we also train a class-conditional linear surrogate of $\text{MUtE}^*_\leftarrow$ via linear regressions to approximate the counterfactual mappings (male $\leftarrow$ female and female $\leftarrow$ male). We benchmark the decoded texts against ground-truth textual counterfactuals—constructed via rule-based substitutions \citep{dearteaga2019bias}—using BLEU and ROUGE (1/2/L $F_1$-scores) to measure lexical overlap, and BERTScore ($F_1$) to quantify dense semantic preservation. Finally, to evaluate the efficacy of the latent intervention, we define the \textit{gender substitution rate} ($r_\text{g}$) as the proportion of generated gender indicators (\textit{he, him, his, mr, m, himself} and \textit{she, her, hers, ms, mrs, herself}) that successfully match the target counterfactual class.

Qualitative examples of the inverted mapping, $\text{MUtE}^*_\leftarrow$ (Table \ref{tab:counterfactual_texts}), demonstrate effective gender substitution with high semantic fidelity. Quantitatively (Table \ref{tab:counterfactual_generation_evaluation}), although texts decoded from $\text{MUtE}^*_\leftarrow$ exhibit lower $n$-gram overlap (BLEU/ROUGE) with the source text compared to baselines, they sustain high BERTScores and achieve a significantly superior gender substitution rate. This discrepancy exposes a fundamental failure mode of standard linear steering baselines: their high lexical overlap is largely an artifact of under-intervention. Operating essentially as near-identity functions, these methods artificially inflate $n$-gram metrics by passively reconstructing the original text. In contrast, by enforcing high-order alignment of the conditional distributions, $\text{MUtE}^*_\leftarrow$ induces deep structural alterations to execute a robust semantic intervention, while preserving the core semantic utility. Notably, the linear surrogate derived from $\text{MUtE}^*_\leftarrow$ achieves comparable efficacy, thereby introducing a novel, computationally efficient linear steering mechanism for targeted concept intervention.

\begin{table}[t]
    \centering
    \caption{Counterfactual texts decoded after intervention on gender via MUtE$^*_\leftarrow$ and Linear OT steering on the \textsc{Bias in Bios} dataset. Original gender markers are \textbf{bolded}. Post-intervention markers that successfully align with the target gender are highlighted in green and bolded, whereas contradictory or unchanged markers are highlighted in red and italic.\\}
    \begin{tabular}{p{0.1\linewidth}p{0.07\linewidth}p{0.75\linewidth}}
        \toprule
        \textbf{Model} &  Gender &  Text\\ 
        \midrule
        \textit{original} & male & \footnotesize \textbf{He} received \textbf{his} BA in Mathematics Education from Mercyhurst College in Eric, Pennsylvania, and \textbf{his} PhD in Mathematics from the University of South Carolina.\\
        
        Linear OT & female & \footnotesize \redhl{He} received \greenhl{her} BA in Mathematics Education from Mercyhurst College in Eric, Pennsylvania, and \greenhl{her} PhD in Mathematics from the University of South Carolina, Tri Carolina.\\

        \rowcolor{black!5}MUtE$^*_\leftarrow$ & female & \footnotesize \greenhl{She} received \greenhl{her} BA in Mathematics Education from Mercyhurst College in South Carolina, and \greenhl{her} PhD in the Department of Mathematics and Education at Eric University in Pennsylvania.\\
        \hline \\[-1.5ex]
        \textit{original} & female & \footnotesize \textbf{She} participated in projects concerning data analysis. \textbf{Her} research interests include applications of game theory to queueing networks, and to inventory management.\\
        
        Linear OT & male & \footnotesize \redhl{She} participated in projects related to data analysis. \greenhl{His} research interests include applications of game theory to inventory management, data management, and queueing networks .\\
        
        \rowcolor{black!5}MUtE$^*_\leftarrow$ & male & \footnotesize \greenhl{He} participated on projects related to data analysis. \greenhl{His} research interests include applications of game theory to inventory management, and stochastic queueing game networks.\\
        \bottomrule
    \end{tabular}
    \label{tab:counterfactual_texts}
\end{table}

\begin{table}[t]
    \centering
    \caption{Counterfactual text generation evaluation. \textit{True CFR} (CounterFactual Representation) corresponds to texts decoded from the embeddings of true counterfactual text.\\}
    \begin{tabular}{lrrrr}
        \toprule
        \textbf{Model} &   \textbf{\scriptsize BLEU} $\uparrow$ &  \textbf{\scriptsize ROUGE} $\uparrow$ & \bm{$r_\text{g}$} $\uparrow$  & \textbf{\scriptsize BERTScore}  $\uparrow$\\ 
        
        \midrule
        \textit{True CFR} & \textit{78.68} & \textit{.92/.80/.88} & \textit{.98} & \textit{0.98}\\
        Mean diff. & 61.00 & .84/.64/.78 & .46 & 0.97\\
        Linear OT & 60.45 & .85/.65/.78 & .65 & 0.97\\
        \rowcolor{black!5}MUtE$^*_\leftarrow$ & 32.93 & .67/.38/.57 & .85 & 0.93\\
        \rowcolor{black!5} 	$\hookrightarrow$ Linear Surrogate & 31.18 & .66/.36/.56 & .92 & 0.93\\
        \bottomrule
    \end{tabular}
    \label{tab:counterfactual_generation_evaluation}
\end{table}

\section{Discussion and Future Directions}

This work builds upon recent frameworks unifying concept erasure and counterfactual representation generation, demonstrating they are fundamentally dual interventions. By formalizing the navigation between erased and counterfactual continuous spaces via diffeomorphisms, our approach establishes a versatile paradigm that readily extends beyond NLP. 

Our approach bypasses gradient-based optimization, yielding an efficient, deterministic training phase in high-dimensional continuous spaces. This training efficiency, however, trades off against inference latency: the forward mapping cost scales linearly with the number of iterations $T$ required for convergence. To circumvent the $\mathcal{O}(T)$ inference bottleneck, the exact multi-step mapping $f$ can be distilled into an amortized surrogate $f_\theta$ by minimizing the regression loss ${\mathcal{L} = \mathbb{E}\left[ \| f(X) - f_\theta(X) \|_2^2 \right]}$. This offline distillation seamlessly accommodates diverse deployment constraints. To strictly preserve the bijectivity required for dual counterfactual generation, $f_\theta$ can be parameterized using class-specific Invertible Neural Networks (INNs) or memory-efficient Conditional INNs (cINNs) \citep{ardizzone2020conditional}. When solely forward erasure is required, relaxing the bijectivity constraint via standard feed-forward networks (e.g., MLPs) maximizes inference throughput.

As a preliminary proof-of-concept, we empirically demonstrate in Appendix \ref{sec:appendix_distillation} that a MLP surrogate successfully preserves both the predictive utility and fairness guarantees of the exact $\text{MUtE}$ representations on downstream tasks. This confirms that a lightweight feed-forward network can adequately approximate the underlying transformations, offering an efficient alternative for practical deployment. We leave extensive evaluations of these distillation strategies to future work.

Beyond resolving computational limitations, subsequent research must expand the framework's causal expressivity. While our current empirical estimator enforces a rigid translational bias, real-world causal factors of variation frequently interact via complex, non-linear mechanisms exhibiting heterogeneous geometric signatures \citep{scholkopf2021toward}. Formulating and integrating expressive geometric priors to capture these diverse, non-linear causal interventions remains an open challenge that necessitates novel theoretical approaches and dedicated future research.

\section{Conclusion}

This work revisits the problem of discrete concept erasure at optimality, formally defined as achieving perfect privacy while maximizing downstream utility preservation. We derive a class of theoretically optimal erasure functions that naturally induce a dual, deterministic counterfactual mapping within the continuous representation space. To bridge theory and practice, we propose a computationally tractable implementation that aligns with this theoretical framework. We demonstrate its empirical efficacy across real-world Natural Language Processing tasks, successfully exploiting the geometric phenomenon that many latent concepts in modern language models manifest as rigid location shifts. Notably, our method proves highly effective for both algorithmic bias mitigation and the generation of counterfactual texts. Future work will explore extending this framework to other data modalities such as image representations in computer vision, and integrating complex geometric priors to model diverse causal interventions.


\section*{Limitations}

Our framework is strictly formulated for the erasure of a single, discrete concept. It does not naturally extend to continuous sensitive attributes—a regime where methods such as FaRM \citep{chowdhury2022learning} and KRaM \citep{chowdhury2024robust} are currently better suited—nor does it directly accommodate the simultaneous joint erasure of multiple intersecting concepts. 

While enforcing a translational geometric prior is highly effective for many latent concepts in modern text encoders, real-world causal factors of variation frequently interact via highly non-linear mechanisms. Consequently, applying MUtE to arbitrarily complex, entangled representations without verifying the underlying topological assumptions risks suboptimal or unpredictable interventions.

\section*{Ethical Considerations}

In this work, we intervene on sensitive demographic attributes—such as gender, race, and religion—by formulating them as discrete categorical variables drawn from a restricted label set. In real-world applications, operationalizing these attributes requires rigorous consensus, a process frequently hindered by heterogeneous cultural, ethical, and legal contexts. Furthermore, the forced erasure of concept-related representations inherently induces a loss of information. This utility degradation can severely impair downstream predictive performance, posing significant risks when deploying these models in high-stakes domains (e.g., healthcare, criminal justice, or resource allocation). Consequently, algorithmic fairness cannot be the sole evaluation criterion in practice. Finally, targeted interventions on specific data dimensions carry unintended systemic risks: erasing a single sensitive attribute may inadvertently incentivize the model to exploit unprotected proxy variables, potentially exacerbating representation biases against intersecting demographic groups.

\bibliography{bibliography}
\bibliographystyle{tmlr}

\include{appendix}

\end{document}

%% file: math_commands.tex
\usepackage{amsmath,amsfonts,bm}

\def\eqref#1{equation~\ref{#1}}

\def\1{\bm{1}}

\DeclareMathAlphabet{\mathsfit}{\encodingdefault}{\sfdefault}{m}{sl}
\SetMathAlphabet{\mathsfit}{bold}{\encodingdefault}{\sfdefault}{bx}{n}

\newcommand{\R}{\mathbb{R}}



%% file: appendix.tex
\appendix

\section{Conditional Negentropy Evolution}
\label{sec:app:noisy_routing}

In this section, we analyse the evolution of the negentropy for each class of samples. We first compute the theoretical negentropy reduction under a perfect routing assumption (Section \ref{app:negentropy_reduction_perfect_routing}). Subsequently, we rigorously evaluate the negentropy evolution under the empirical noisy routing assumption (Section \ref{app:negentropy_reduction_noisy_routing}).

Let $i \in \mathcal{Z}$ denote a target concept class and $t$ denote an arbitrary iteration step of MUtE. Let $R^{(t)}$ and $\{\psi_k^{(t)}\}_{k\in\mathcal{Z}}$ be the orthogonal rotation matrix and the class-conditional marginal Gaussianizations at step $t$, respectively. For notational brevity, we omit the iteration superscript $(t)$ where unambiguous.

Let $X_i^{(t)}$ be a random variable distributed according to the true class-conditional probability density function (PDF) $p_i^{(t)}$. We define the rotated variable as $\tilde{X}_i = R X_i^{(t)}$, with corresponding joint PDF $\tilde{p}_i^{(t)}$ and marginal PDFs $\tilde{p}_{i,d}^{(t)}$ for each dimension $d \in \{1, \dots, D\}$. 

We denote the rotated decision regions, partitioned by the potentially noisy empirical predictor $\eta^*$, as $\tilde{\Omega}_k = \{ \tilde{x} \in \mathbb{R}^D \mid \eta^*(R^{-1}\tilde{x}) = k \}$. Using the indicator function $\mathbb{I}_{\tilde{\Omega}_k}$, the piecewise forward mapping of a sample $x \sim X_i^{(t)}$ (Equation \ref{eq:SIG}) can be expressed over the rotated space as:
$$ x^{(t+1)} = \psi_{\eta^*(x)} (R x) = \sum_{k \in \mathcal{Z}} \mathbb{I}_{\tilde{\Omega}_k}(\tilde{x}) \psi_k(\tilde{x}) $$

Consequently, the pushforward measure for the entire class $i$ is given by:
\begin{equation}
\label{app:eq:X_i_plus_one}
X_i^{(t+1)} = \sum_{k \in \mathcal{Z}} \mathbb{I}_{\tilde{\Omega}_k}(\tilde{X}_i) \psi_k(\tilde{X}_i)
\end{equation}

We track the convergence using the negentropy $J(X) = D_{\mathrm{KL}}(P \parallel \mathcal{N}(0, I))$, which can be decomposed into the differential entropy and a cross-entropy penalty against the standard normal PDF $\phi$ (which is the derivative of the standard normal CDF $\Phi'(x) = \phi(x)$):
\begin{equation}
\label{app:eq:negentropy_entropy}
J(X) = -h(X) - \mathbb{E}_{x \sim X} \left[ \log \prod_{d=1}^D \phi(x_d) \right] 
\end{equation}
where the differential entropy is defined as:
\begin{equation}
\label{app:eq:entropy}
h(X) = - \mathbb{E}_{x \sim X} [\log p(x)] 
\end{equation}

\subsection{Negentropy Reduction Under Perfect Routing}
\label{app:negentropy_reduction_perfect_routing}

Under the assumption of an optimal oracle predictor ($\eta^*(x) = i$), all samples belonging to class $i$ are transformed strictly via the corresponding diffeomorphism $\psi_i \circ R$. Following standard Iterative Gaussianization (RBIG) principles \citep{laparra2011iterative}, each transformation strictly reduces the negentropy of the distribution. The negentropy reduction under perfect routing is:
$$ \Delta J^*_i = J(X_i^{(t)}) - J((\psi_i \circ R)(X_i^{(t)})) $$

Because orthogonal rotations are entropy-preserving isometries, $J(X_i^{(t)}) = J(\tilde{X}_i)$. Thus:
$$ \Delta J^*_i = J(\tilde{X}_i) - J(\psi_i(\tilde{X}_i)) $$

Recalling property 3.1 from \citet{laparra2011iterative}, this reduction exactly equals the sum of the marginal negentropies of the rotated representations:
$$ \Delta J^*_i = \sum_{d=1}^D \mathbb{E}_{\tilde{x} \sim \tilde{X}_i} \left[ \log \frac{\tilde{p}_{i,d}^{(t)}(\tilde{x}_d)}{\phi(\tilde{x}_d)} \right] \geq 0 $$

\subsection{Negentropy Evolution Under Noisy Routing}
\label{app:negentropy_reduction_noisy_routing}

We now evaluate the evolution of the negentropy $\Delta J_i$ under a realistic noisy routing regime, where overlaps in conditional distributions yield an irreducible Bayes error. The exact negentropy reduction can be decomposed as:
\begin{equation}
\label{app:eq:negentropy_evolution}
\begin{aligned}
\Delta J_i &= J(X_i^{(t)}) - J(X_i^{(t+1)}) \\
&= \Delta J^*_i - \underbrace{\left( J(X_i^{(t+1)}) - J(\psi_i(\tilde{X}_i)) \right)}_{:=E_i}
\end{aligned}
\end{equation}

To explicitly compute $J(X_i^{(t+1)})$, we first evaluate the differential entropy $h(X_i^{(t+1)})$ of the shattered pushforward measure (Equation \ref{app:eq:negentropy_entropy}). To rigorously account for the non-differentiable decision boundaries $\partial \tilde{\Omega}_k$, we decompose the integral over the latent space into a piecewise sum over the disjoint regions $\tilde{\Omega}_k$. Because the boundaries possess a Lebesgue measure of zero, they do not contribute to the integral:
\begin{equation*}
h(X_i^{(t+1)}) = -\sum_{k \in \mathcal{Z}} \int_{\tilde{\Omega}_k} p_i^{(t+1)}(y) \log p_i^{(t+1)}(y) dy 
\end{equation*}

Applying the change of variables $y = \psi_k(\tilde{x})$ locally within the interior of each region $\tilde{\Omega}_k$ yields:
\begin{equation}
\label{app:eq:entropy_variable_change}
\begin{array}{ll}
h(X_i^{(t+1)}) & = -\sum_{k \in \mathcal{Z}} \int_{\tilde{\Omega}_k} \tilde{p}_i^{(t)}(\tilde{x}) \log \left( \frac{\tilde{p}_i^{(t)}(\tilde{x})}{\vert \det \mathbf{J}_{\psi_k}(\tilde{x}) \vert} \right) d\tilde{x} \\
& = h(\tilde{X}_i) + \sum_{k \in \mathcal{Z}} \int_{\tilde{\Omega}_k} \tilde{p}_i^{(t)}(\tilde{x}) \log \left( \vert \det \mathbf{J}_{\psi_k}(\tilde{x}) \vert \right) d\tilde{x}
\end{array}
\end{equation}

Because the marginal transformation $\psi_k$ operates independently across dimensions, its Jacobian matrix $\mathbf{J}_{\psi_k}$ is purely diagonal:
$$ \det \mathbf{J}_{\psi_k}(\tilde{x}) = \prod_{d=1}^D \frac{d}{d\tilde{x}_d} \psi_{k, d}(\tilde{x}_d) $$

Applying the chain rule and the inverse function theorem to the marginal uniformization ${\psi_{k, d}(\tilde{x}_d) = \Phi^{-1} \left( \int_{-\infty}^{\tilde{x}_d} p_{k,d}^{(t)}(u)du \right)}$ (Equation \ref{eq:conditional_marginal_gaussianization}), we have:

\begin{equation*}
\frac{d}{d\tilde{x}_d} \psi_{k, d}(\tilde{x}_d) = \frac{\tilde{p}_{k,d}^{(t)}(\tilde{x}_d)}{\phi(\psi_{k, d}(\tilde{x}_d))}
\end{equation*}

And thus:

\begin{equation}
\label{app:eq:abs_log_det}
\vert \det \mathbf{J}_{\psi_k}(\tilde{x}) \vert = \prod_{d=1}^D \frac{\tilde{p}_{k,d}^{(t)}(\tilde{x}_d)}{\phi(\psi_{k, d}(\tilde{x}_d))}
\end{equation}

Substituting Equation \ref{app:eq:abs_log_det} back into the differential entropy expansion (Equation \ref{app:eq:entropy_variable_change}), and aggregating the piecewise integrals via the empirical predictor $\eta^*(x)$, we obtain:
\begin{equation}
\label{app:eq:entropy_developped}
h(X_i^{(t+1)}) = h(\tilde{X}_i) 
+ \mathbb{E}_{\tilde{x} \sim \tilde{X}_i} \left[ \log \prod_{d=1}^D \tilde{p}_{\eta^*(x),d}^{(t)}(\tilde{x}_d) \right] 
- \mathbb{E}_{\tilde{x} \sim \tilde{X}_i} \left[ \log \prod_{d=1}^D \phi(\psi_{\eta^*(x), d}(\tilde{x}_d)) \right] 
\end{equation}

We formulate the negentropy of the pushforward measure $J(X_i^{(t+1)})$ by substituting Equation \ref{app:eq:entropy_developped} into Equation \ref{app:eq:negentropy_entropy}. Observing that the standard normal cross-entropy terms exactly cancel out, the expression simplifies to:
\begin{equation}
\label{app:eq:negentropy_final}
J(X_i^{(t+1)}) = -h(\tilde{X}_i) - \mathbb{E}_{\tilde{x} \sim \tilde{X}_i} \left[ \log \prod_{d=1}^D \tilde{p}_{\eta^*(x),d}^{(t)}(\tilde{x}_d) \right]
\end{equation}

By corollary, the negentropy term corresponding to a perfect oracle routing ($J(\psi_i(\tilde{X}_i))$) is recovered by uniformly substituting $\eta^*(x) = i$:
\begin{equation}
\label{app:eq:negentropy_compare}
J(\psi_i(\tilde{X}_i)) = -h(\tilde{X}_i) - \mathbb{E}_{\tilde{x} \sim \tilde{X}_i} \left[ \log \prod_{d=1}^D \tilde{p}_{i,d}^{(t)}(\tilde{x}_d) \right]
\end{equation}

Subtracting Equation \ref{app:eq:negentropy_compare} from Equation \ref{app:eq:negentropy_final} isolates the routing penalty $E_i$ introduced in Equation \ref{app:eq:negentropy_evolution}:

\begin{equation}
\label{app:eq:routing_error}
E_i = \mathbb{E}_{\tilde{x} \sim \tilde{X}_i} \left[ \log \prod_{d=1}^D \frac{\tilde{p}_{i,d}^{(t)}(\tilde{x}_d)}{\tilde{p}_{\eta^*(x),d}^{(t)}(\tilde{x}_d)} \right] 
\end{equation}

It needs to be noted that $E_i$ is entirely dictated by the misclassified regions, and rewrites as:

\begin{equation}
\label{app:eq:routing_error_misclassified}
E_i = \sum_{k \neq i} \int_{\tilde{\Omega}_k} \tilde{p}_i^{(t)}(\tilde{x}) \log \prod_{d=1}^D \frac{\tilde{p}_{i,d}^{(t)}(\tilde{x}_d)}{\tilde{p}_{k,d}^{(t)}(\tilde{x}_d)} \, \mathrm{d}\tilde{x}
\end{equation}

The term $E_i$ represents the exact entropic penalty incurred by spatial tearing. It mathematically quantifies the cross-entropy mismatch caused by evaluating the true marginal density $\tilde{p}_{i,d}^{(t)}$ using the estimators of the mispredicted class $\tilde{p}_{\eta^*(x),d}^{(t)}$. Because the decision boundaries $\partial \tilde{\Omega}_k$ are non-differentiable, samples routed incorrectly inevitably inject non-Gaussian artifacts into the pushforward measure. Consequently, strict monotonic negentropy reduction ($\Delta J_i > 0$) is not guaranteed. Noisy routing injects entropy ($E_i > 0$), while the subsequent rotation and exact marginalization forcibly remove it ($\Delta J^*_i > 0$).

Crucially, Equation \ref{app:eq:routing_error_misclassified} demonstrates that the penalty $E_i$ diverges to infinity if a mispredicted marginal density $\tilde{p}_{k,d}^{(t)}$ evaluates to zero over regions where the true density $\tilde{p}_{i,d}^{(t)}$ is strictly positive. To rigorously constrain this divergence in practice, we enforce a strict minimum threshold $\alpha > 0$ on all empirical marginal density estimators. By bounding the denominator, the maximum log-ratio across all $D$ dimensions becomes strictly finite. Defining $\epsilon_i = \sum_{k \neq i} \int_{\tilde{\Omega}_k} \tilde{p}_i^{(t)}(\tilde{x}) \, \mathrm{d}\tilde{x} \equiv \mathbf{P}(\eta^*(\tilde{X}_i) \neq i)$ as the class-specific empirical routing error, we can factor it out of the domain of integration, yielding an upper bound of:

\begin{equation}
\label{app:eq:routing_error_bound}
    E_i \leq \mathcal{O}(-\epsilon_i D \log \alpha)
\end{equation}

This thresholding strategy guarantees a strictly finite supremum for the spatial tearing penalty, explicitly tying the worst-case algorithmic instability to the intrinsic Bayes error of the latent space.

\section{Datasets}
\label{sec:app:datasets}

\paragraph{\textsc{GloVe}} 
This dataset is a subset of the 150,000 most frequent word embeddings from the original \textsc{GloVe} corpus \cite{pennington2014glove}. Words are categorized into three discrete concept classes (male-biased, female-biased, and neutral) based on the magnitude of their projection onto the gender direction. This direction is formally defined as the principal component of the subspace spanned by gendered word-pair differences. The dataset comprises 21,996 embeddings, partitioned into 10,777 for training (49\%), 4,620 for validation (21\%), and 6,599 for testing (30\%).

\paragraph{\textsc{Bias in Bios}} 
This is a real-world benchmark of short biographies scraped from the web, designed specifically to study gender bias in NLP \cite{dearteaga2019bias}. Each sample is annotated with a binary gender attribute and one of 28 occupation labels. We utilize the subset curated by \cite{ravfogel2020null}, which preserves $\sim$98\% of the original corpus. The dataset exhibits severe historical gender-occupation correlations. It contains 399,423 biographies, stratified by occupation into 255,710 training (64\%), 39,369 validation (10\%), and 98,344 testing (25\%) samples.

\paragraph{\textsc{DIAL}} 
Derived from the DeepMoji corpus \cite{blodgett2016demographic}, \textsc{DIAL} is a Twitter-based sentiment classification benchmark. Each observation includes a binary downstream sentiment label (\textit{happy} or \textit{sad}) and a binary demographic attribute corresponding to the linguistic dialect: African-American English (AAE) or Standard American English (SAE). We adopt the version pre-processed by \cite{chowdhury2024robust}, which guarantees perfect balance across both race and sentiment labels. The dataset totals 175,996 samples, split into 160,000 for training (91\%), 8,000 for validation (4.5\%), and 7,996 for testing (4.5\%).

\paragraph{\textsc{Jigsaw}} 
Based on the Jigsaw Toxicity Classification benchmark, this dataset is used for the downstream task of binary toxicity detection. We target religion as the sensitive concept, operationalized as a discrete categorical variable with five labels: \textit{Buddhist}, \textit{Christian}, \textit{Hindu}, \textit{Jewish}, and \textit{Muslim}. The corpus comprises 96,492 samples, divided into 87,434 for training (91\%) and 9,058 for testing (9\%).

\section{Training}
\label{sec:app:training-parameters}

\subsection{Hardware and Environment}

All experiments are conducted on an NVIDIA GeForce RTX 2080 Ti GPU with 11GB of VRAM and an Intel(R) Core(TM) i9-9900K CPU, using PyTorch 2.8.0 and Python 3.12.9. We employ CUDA 12.6 for accelerated computations.

\subsection{MUtE}

We selected $T$ via early stopping, halting the procedure when the validation probe accuracy plateaued at the majority-class baseline, resulting in: $T=100$ for \textsc{GloVe}, $T=70$ for \textsc{Bias in Bios}, $T=150$ for \textsc{DIAL} and \textsc{Jigsaw}. 

\subsection{Other models settings}

\paragraph{LEACE.} We relied on the publicly available implementation from: \newline \url{https://github.com/EleutherAI/concept-erasure}.

\paragraph{KRaM and FaRM.} 
We reimplement both KRaM \cite{chowdhury2024robust} and FaRM \cite{chowdhury2022learning} following a standardized training protocol. For both architectures, the hidden dimension at each layer is strictly constrained to match the input feature dimension. Network depth is task-dependent: we utilize 4 layers for \textsc{GloVe}, \textsc{Bias in Bios}, and \textsc{Jigsaw}, and 7 layers for \textsc{DIAL}. The networks are optimized over 50 epochs with a batch size of 512, utilizing a learning rate of $10^{-3}$ and a weight decay coefficient of $10^{-5}$. For the KRaM objective specifically, the regularization weighting hyperparameter is set to $\lambda = 0.7$.

\paragraph{$\overline{\mathrm{\textbf{L}}}$EOPARD.}
We reimplemented the $\overline{\mathrm{L}}$EOPARD framework \cite{saillenfest2025nonlinear}. For the \textsc{GloVe}, \textsc{Bias in bios}, and \textsc{DIAL} datasets, we adopted the exact hyperparameter configurations recommended by the original authors. Lacking explicit guidelines for \textsc{Jigsaw}, we applied the \textsc{DIAL} configuration to this dataset. Cascaded training was employed across all settings. Specifically, models were trained for $1000$ epochs on \textsc{GloVe}, $100$ epochs on \textsc{Bias in bios}, and $200$ epochs on both \textsc{DIAL} and \textsc{Jigsaw}, utilizing batch sizes of $10777$, $8192$, and $2048$, respectively. The regularization parameter $\gamma$ was set to $200$ for \textsc{GloVe} and $100$ for all other datasets. The initial learning rate was configured to $1.0 \times 10^{-3}$ for \textsc{GloVe} and $5.0 \times 10^{-4}$ for the remaining datasets, subject to a step decay factor of $0.1$ at the training midpoint.

\paragraph{TaCo.} 
We reimplemented the TaCo framework \cite{jourdan2023taco}. Standard TaCo operates as a constrained nonlinear erasure method, optimizing post-erasure representations by ranking removal directions according to their joint importance to both the target concept and a predefined downstream label. For a fair evaluation against unconstrained baselines, we ablated this downstream dependency. Specifically, we adapted TaCo to rank directions based exclusively on their relevance to the target concept, quantified via variance-based sensitivity analysis using Sobol indices. Consistent with \citet{jourdan2023taco}, we initially apply PCA to project the representations onto a $100$-dimensional subspace. To trace the Pareto front (Figure~\ref{fig:utility_privacy_tradeoff}), we sweep the number of filtered dimensions from $5$ to $95$ by step of $5$.

\subsection{Evaluation Protocol}

To evaluate representation quality, both the probing classifiers and the downstream task models are implemented via \texttt{scikit-learn}'s \texttt{MLPClassifier}. The networks are optimized using a constant learning rate of $10^{-4}$ for a maximum of 20 training epochs, which empirically proved sufficient for convergence. To ensure statistical reliability, all reported accuracy and fairness metrics are averaged across five independent experimental runs.

\section{Fairness Metrics}
\label{sec:app:fairness_metrics}

$\mathrm{TPR}^{\mathrm{RMS}}$ is the root mean square (RMS) of the sum of the bias quantified by computing the difference (Gap) in the true positive rate (TPR) of the classifier between individuals with different concept class labels. Formally, for a binary concept and downstream labels $Y$ sampled from the set of downstream labels $\mathcal{Y}$:

\begin{equation}
\begin{split}
    \mathrm{TPR}_{0,y} = p(\hat{Y} = y | Z = 0, Y = y) \\
    \mathrm{Gap}_{y} = \mathrm{TPR}_{1,y} - \mathrm{TPR}_{0,y} \\
    \mathrm{TPR}^{\mathrm{RMS}} =  \sqrt{\frac{1}{|\mathcal{Y}|}\sum_{y\in\mathcal{Y}} (\mathrm{Gap}_{y})^2 }
\end{split}
\end{equation}

Intuitively, the true-positive-rate of a “fair” classifier should not be sensitive to the values of the protected attributes. 

For completeness, we also report demographic parity which measures the difference in prediction w.r.t. to a protected attribute and is achieved when the probability of a certain prediction is not dependent on sensitive group membership:

\begin{equation}
    \mathrm{DP} = \sum_{y\in \mathcal{Y}} |p(\hat{Y}=y|Z=0) - p(\hat{Y}=y|Z=1)|\\
\end{equation}

\begin{table*}[t!]
    \centering
    \small 
    \begin{tabular}{lrrrrr}
        \toprule
        \textbf{Model} & \textbf{Inference time} & \bm{$a_y$} (\%) $\uparrow$ & \bm{$\mathrm{TPR}^{\mathrm{RMS}}$} $\downarrow$ & \textbf{DP} $\downarrow$ & \bm{$a_z$} (\%) \\ 
        
        \midrule
        \multicolumn{6}{l}{\textsc{Bias in Bios}} \\
        MUtE          & 18.7s & 70.2{\scriptsize $\;\pm\;$0.2} & 0.091{\scriptsize $\;\pm\;$0.002} & 0.391{\scriptsize $\;\pm\;$0.004} & 53.2{\scriptsize $\;\pm\;$0.4} \\
        MLP surrogate & 2.1s  & 71.9{\scriptsize $\;\pm\;$0.2} & 0.089{\scriptsize $\;\pm\;$0.005} & 0.413{\scriptsize $\;\pm\;$0.005} & 53.3{\scriptsize $\;\pm\;$0.2} \\

        \midrule
        \multicolumn{6}{l}{\textsc{DIAL}} \\
        MUtE          & 14.0s & 71.0{\scriptsize $\;\pm\;$0.3} & 0.094{\scriptsize $\;\pm\;$0.005} & 0.086{\scriptsize $\;\pm\;$0.012} & 50.2{\scriptsize $\;\pm\;$0.4} \\
        MLP surrogate & 0.1s  & 71.4{\scriptsize $\;\pm\;$0.1} & 0.091{\scriptsize $\;\pm\;$0.006}& 0.092{\scriptsize $\;\pm\;$0.016} & 50.0{\scriptsize $\;\pm\;$0.4} \\
        \bottomrule
    \end{tabular}
    \caption{Empirical evaluation of the MLP surrogate's fidelity relative to the exact $\text{MUtE}$ mapping. Inference time indicates the total duration required to map the respective test sets. Downstream performance and fairness metrics are evaluated using classifiers trained on the exact $\text{MUtE}$ representations but tested on representations generated by the designated model at inference time.}
    \label{tab:surrogate_model_results}
\end{table*}

\section{Distillation of the MUtE Mapping}
\label{sec:appendix_distillation}

As a preliminary proof-of-concept for offline distillation, we approximate a pre-trained exact $\text{MUtE}$ mapping using a Multi-Layer Perceptron (MLP) surrogate. The surrogate pipeline comprises an initial Principal Component Analysis (PCA) whitening step (fitted on the training distribution) followed by a 4-layer MLP. To ensure sufficient capacity, the hidden dimension is set to $2d$, where $d$ represents the input feature dimension. Intermediate layers consist of a linear transformation, Layer Normalization, and a $\tanh$ activation, culminating in a linear output layer. 

We train the surrogate model to minimize the Mean Squared Error (MSE) between its outputs and the exact $\text{MUtE}$ representations. Optimization is performed using AdamW with a learning rate of $10^{-3}$ and a batch size of $2,048$ for a maximum of $1,500$ epochs, holding out $2,000$ samples for validation monitoring.

As detailed in Table \ref{tab:surrogate_model_results}, the MLP surrogate strictly preserves both the predictive utility ($a_y$) and the fairness guarantees ($\mathrm{TPR}^{\mathrm{RMS}}$, DP, $a_z$) of the exact $\text{MUtE}$ mapping. Crucially, the surrogate accelerates inference by at least an order of magnitude. This empirically validates that a lightweight feed-forward network can successfully capture the underlying geometric manifold of the optimal erasure function, providing an efficient alternative for latency-sensitive downstream deployments.